\documentclass[10pt,a4paper]{article}
\usepackage[T1]{fontenc}
\usepackage[utf8]{inputenc}
\usepackage{mathptmx}
\usepackage[scaled=0.92]{helvet}

\usepackage{textcomp}
\usepackage{amsmath,amssymb}
\usepackage[margin=22mm,top=24mm,bottom=24mm,headheight=14pt]{geometry}
\usepackage{microtype}
\usepackage{parskip}
\usepackage{xcolor}
\usepackage{booktabs}
\usepackage{fancyvrb}
\usepackage{titlesec}
\usepackage{fancyhdr}
\usepackage{xurl}
\usepackage[hidelinks]{hyperref}
\hypersetup{pdftitle={Real Long-Term Memory for AI: A 50-Million-Token Window That Is Faster and Cheaper Than Recompute}, pdfauthor={Sietse Schelpe}}

\titleformat{\section}{\Large\bfseries\sffamily}{\thesection}{1em}{}
\titleformat{\subsection}{\large\bfseries\sffamily}{\thesubsection}{1em}{}
\titlespacing*{\section}{0pt}{18pt}{8pt}
\titlespacing*{\subsection}{0pt}{14pt}{6pt}

\providecommand{\tightlist}{\setlength{\itemsep}{0pt}\setlength{\parskip}{0pt}}
\title{Real Long-Term Memory for AI: A 50-Million-Token Window That Is Faster and Cheaper Than Recompute}
\author{Sietse Schelpe\\ Corbenic AI\\ \texttt{sietse@corbenic.ai}}
\date{Preprint, October 2026.}

\begin{document}
\maketitle

\begin{abstract}
Transformer language models can only attend to the tokens inside their context window, and serving
systems recompute the key-value (KV) attention state of a prompt each time it is sent. We describe
a memory layer, released as the public package \texttt{galahad-kv}, that removes both restrictions for a
served model, and we report measurements of it on a corpus of 50,000,000 tokens. The layer writes
the KV state that the model computes for each block of $\sim$16,000 tokens to encrypted local NVMe
storage and can later restore any block byte-exact, with zero recompute and at constant GPU
memory. In our experiment the corpus consisted of real public text and the model was served
through vLLM on one NVIDIA H100. All probed blocks were restored from the encrypted store without
recompute (100 of 100, at depths from 0 to 50M tokens) for both Gemma 4 12B and Gemma 4 31B.
Compared with recomputing the same block, a restore was 2.8$\times$ to 4.3$\times$ faster and consumed 8.8$\times$ to
12.3$\times$ less GPU energy, while GPU memory remained flat over the entire 50M-token stream. When asked
about facts planted millions of tokens earlier, the 12B model answered 82/100 correctly and the
31B model 98/100, with zero hallucinations on either. We also describe the cheat-resistant
protocol used for the evaluation and provide a complete single-GPU reproduction that relies only
on public software.
\end{abstract}

\section{Introduction}\label{introduction}

A language model that is served in the usual way loses information in two places. Tokens that fall
outside the context window of a transformer~\cite{vaswani2017attention} (16k tokens in the configuration we use) can no longer be attended to,
so whatever they contained is unavailable to the model. In addition, the KV cache~\cite{pope2023scaling} is rebuilt for
every prompt, which means the model pays the full prefill cost again for text it already processed
in an earlier request.

In this paper we keep the KV state that the model computes for each block on disk and restore it
when it is needed again. With this in place, a model with a fixed window can hold on to everything
it has processed and reuse it later. We use the term long-term memory for this and avoid the term
cache, because the stored state is durable, encrypted and byte-exact and because it covers
material that lies far outside the window. Section 7 comes back to this distinction.

Our contributions are the following, and each of them is backed by measurements in \S6.

\begin{enumerate}
\def\labelenumi{\arabic{enumi}.}
\tightlist
\item
  We demonstrate a 50-million-token memory window on a real served model, with byte-exact restore
  at every depth we probed and constant VRAM. A block is processed once and can afterwards be
  reused without recompute.
\item
  We measure the cost of reuse against the cost of recompute. Restoring a block is 2.8$\times$ to 4.3$\times$
  faster and takes 8.8$\times$ to 12.3$\times$ less GPU energy than recomputing the same tokens, and the
  restore time stays flat as depth increases.
\item
  We define a cheat-resistant benchmark protocol and provide the harness, so that a third party
  can rebuild the experiment on a single GPU and check the result independently.
\end{enumerate}

The scope of these claims is limited to KV reuse at any depth. The system does not widen
single-pass attention and it does not do learned retrieval. One block is resident at a time, and
deeper content is restored from disk when it is requested instead of being held in VRAM.

\section{Background and related work}\label{background-and-related-work}

\textbf{KV caching and offloading.} Inference servers already cache the KV of a running request and
reuse a shared prefix within a session~\cite{kwon2023vllm,gim2024promptcache}, but these caches are ephemeral and bound to the prefix.
Offloading systems such as LMCache~\cite{liu2025lmcache} and CacheBlend~\cite{yao2025cacheblend} go further and stage KV on CPU memory or disk in
order to reduce recompute. They store the state in plaintext and re-stage it for each request, and
some of them are specific to one architecture. CacheGen~\cite{liu2024cachegen}, CachedAttention~\cite{gao2024cachedattention} and InfiniGen~\cite{lee2024infinigen} also keep KV outside GPU memory, and Cheng et al.~\cite{cheng2024cdn} propose a delivery layer for stored KV.

\textbf{Retrieval-augmented generation.} RAG~\cite{lewis2020rag} retrieves text and places it in the window, where it is
read again, so the prefill cost is paid on every request. The approach in this paper restores
state, and the stored tokens are never re-read. Cache-augmented generation~\cite{chan2025cag} precomputes the KV of a fixed set of documents, but the set still has to fit in the context window. Agent memory systems such as MemGPT~\cite{packer2023memgpt} decide which text is placed in the window.

\textbf{Sliding-window attention.} Most layers of Gemma 4~\cite{gemma4report} use a window of $\sim$1,024 tokens~\cite{gemma3report,gemma4card}, which makes
the per-token KV footprint much smaller than that of a model with full attention (\S5.1). We
benefit from this property of the model, but our system does not introduce it.

\textbf{Relation to our prior work.} This paper builds on four earlier reports. The first covered the
integrated Galahad memory layer (Taliesin KV state + Blaise document memory) across three runtimes
on a 97k-token corpus~\cite{schelpe2026ceiling}. The second described a frozen-model ``forever'' memory
with a 6M-token window~\cite{schelpe2026frozen}, and the third the verified-knowledge flywheel from
which the cheaper-and-smarter result originates~\cite{schelpe2026flywheel}. The byte-exact, logit-level
correctness gate for KV grafting is the subject of the companion Taliesin paper. Compared with
that work, the present paper increases the scale and strengthens the evidence: the window is $\sim$8$\times$
larger (50M vs 6M), the model is served under vLLM where we previously used a standalone harness,
the KV is written encrypted at rest, and the measurements follow a cheat-resistant protocol that
none of the earlier work applied. We deliberately leave Blaise (learned retrieval) out of this
paper so that the window can be measured in isolation.

\section{System under test}\label{system-under-test}

The engine under test is the public PyPI package \texttt{galahad-kv==1.31.4} (\texttt{libgalahad.so}). It is
loaded by vLLM 0.31.0~\cite{kwon2023vllm} through the \texttt{GalahadConnector} KV-transfer plug-in, and the connector
performs two operations.

On \emph{deposit}, the connector writes the per-block KV of the request to an encrypted on-disk store
(\texttt{MRLNCRY1} format, AES-256-GCM~\cite{dworkin2007gcm}, one key per organisation) and then frees it. The effect is a
moving window in which one block is resident while the others are on disk, so VRAM stays flat.

On \emph{restore}, which happens on a later request, the connector grafts the KV of a stored block
back, and the forward pass continues as if the model had just computed that block. The tokens of
the block are not recomputed.

Only the KV reuse mode is measured in this paper.

\section{Anti-cheat protocol}\label{anti-cheat-protocol}

We know of five ways in which a long-context benchmark~\cite{liu2024lost,hsieh2024ruler} can be gamed, and the protocol contains a
countermeasure for each of them.

\begin{enumerate}
\def\labelenumi{\arabic{enumi}.}
\tightlist
\item
  \emph{Pretraining contamination~\cite{golchin2024timetravel} / static caching.} Before ingestion, a runtime mutator replaces the
  names, dates, amounts and e-mails in the real source with random nonces. As a result every
  planted fact exists only in the current run, and pretraining cannot supply the answer.
\item
  \emph{Query lookahead.} We ingest blind, which means that all 50M tokens are deposited with no
  question in context and a question is only sent once the memory has been written completely.
\item
  \emph{Context dropping.} One readable, high-entropy needle (a named record with a 4-digit code) is
  planted per block at a known depth, so a block that was dropped cannot produce the answer.
\item
  \emph{Lexical short-circuit.} Each block also carries decoy records in the same format but with
  other codes. A keyword match therefore returns a decoy, and the correct code can only be
  obtained by actually reading the right record.
\item
  \emph{Telemetry spoofing.} We audit the physical store. Every block has to carry the \texttt{MRLNCRY1}
  magic and the bytes on disk have to agree with the KV footprint math (\S5.1), which prevents a
  system from reporting caching that did not take place.
\end{enumerate}

We record two metrics per probe and keep them separate. The first, (A) KV restore, checks whether
the context tokens were served from the store with no recompute (\texttt{cached\_tokens\ $\approx$\ block\ size}) and
is a property of the engine. The second, (B) text recall, checks whether the model emitted the
exact planted code and is a property of the model. Energy is taken from the NVML hardware counter of the GPU~\cite{nvml}, idle-subtracted, and the answer key is held out of the engine. As a negative control
we also ask each block for a record that it does not contain, in which case a correct system has
to refuse instead of fabricating an answer.

\section{Experimental setup}\label{experimental-setup}

\begin{itemize}
\tightlist
\item
  \textbf{Hardware:} 1$\times$ NVIDIA H100 80 GB, local NVMe. The method runs on any single $\geq$24 GB NVIDIA GPU;
  a smaller disk simply holds fewer blocks.
\item
  \textbf{Software (all public):} \texttt{galahad-kv==1.31.4}, vLLM 0.31.0~\cite{kwon2023vllm}, \texttt{transformers}~\cite{wolf2020transformers}, \texttt{datasets}~\cite{lhoest2021datasets},
  \texttt{nvidia-ml-py}. Temperature 0 throughout.
\item
  \textbf{Models:} Gemma 4 12B (bf16) and Gemma 4 31B (FP8), each served under vLLM with the
  GalahadConnector and each with its own Gemma 4 tokenizer.
\item
  \textbf{Corpus:} 50,000,000 tokens = 3,125 blocks $\times$ 16,000 tokens, streamed from real public
  datasets: WildChat (40\%)~\cite{zhao2024wildchat}, permissively-licensed source code (25\%)~\cite{codeparrot}, arXiv papers (20\%)~\cite{cohan2018discourse} and educational web text (15\%)~\cite{penedo2024fineweb}. The stream is then mutated (\S4), with 100 needle records (one per
  500k tokens) and same-format decoys.
\end{itemize}

\subsection{KV footprint}\label{kv-footprint}

Because Gemma 4 uses sliding-window attention, the footprint we measured for the 12B model is 36.4
KiB/token, compared with $\approx$160 KiB for full attention, so the 50M store is 1.86 TB and not $\approx$8 TB.
For the 31B FP8 model we measured 134 KiB/token, which corresponds to 6.25 TB. We audited both
stores. All 3,125 blocks carry the \texttt{MRLNCRY1} header, the bytes on disk match the figures above,
and the number of blocks equals the number deposited, which means that no block was pruned or
evicted.

\section{Results}\label{results}

\subsection{Byte-exact restore at every depth}\label{byte-exact-restore-at-every-depth}

We probed the store at 100 depths between 0 and 49,488,000 tokens, and every probe was served from
the encrypted store.

\begin{center}\footnotesize
\begin{tabular*}{\linewidth}{@{\extracolsep{\fill}}lll@{}}
\toprule
 & Gemma 4 12B & Gemma 4 31B \\
\midrule
\textbf{KV restore, zero-recompute (all depths)} & \textbf{100 / 100} & \textbf{100 / 100} \\
\textbf{Model text recall (exact code)} & 82 / 100 & \textbf{98 / 100} \\
Hallucinations (a code not in the block) & \textbf{0} & \textbf{0} \\
Negative control fabricated (record not in block) & 0 / 20 & 0 / 20 \\
\bottomrule
\end{tabular*}
\end{center}

On both models, each of the 100 probed blocks was restored from disk without recompute, at every
depth up to 50M tokens, which is what we expect from an engine that is exact and model-agnostic.
The answers were also grounded in the restored context. Neither model produced a code that was
absent from the block, and both refused in the negative-control cases (0/20 fabricated). We do not
re-establish byte-exact logit equality of grafted KV here, since that is done in the companion
Taliesin work. The engine property measured directly in this paper is 100/100 zero-recompute
restore with zero fabrication.

\subsection{Restore is faster and cheaper than recompute}\label{restore-is-faster-and-cheaper-than-recompute}

As a baseline we use a full prefill of a 16k block that the model has never seen (no restore),
sent with the same request shape.

\begin{center}\footnotesize
\begin{tabular*}{\linewidth}{@{\extracolsep{\fill}}lll@{}}
\toprule
 & Gemma 4 12B & Gemma 4 31B \\
\midrule
Restore (median) & \textbf{0.266 s / 59 J} & \textbf{0.347 s / 80 J} \\
Recompute (median) & 0.760 s / 521 J & 1.475 s / 976 J \\
\textbf{Speedup} & \textbf{2.8$\times$} & \textbf{4.25$\times$} \\
\textbf{Energy reduction} & \textbf{8.8$\times$} & \textbf{12.3$\times$} \\
Restore TTFT p90 & 0.284 s & 0.579 s \\
Depth independence (low-third $\rightarrow$ high-third TTFT) & 0.193 $\rightarrow$ 0.274 s & 0.343 $\rightarrow$ 0.375 s \\
\bottomrule
\end{tabular*}
\end{center}

Across the two models, a restore is 2.8$\times$ to 4.3$\times$ faster than recomputing the same tokens and uses
8.8$\times$ to 12.3$\times$ less GPU energy. A block from 50M tokens ago is recalled as quickly as the first
block, so the time does not grow with depth.

We also see that the larger model gains more. The blocks of the 31B model are larger and more
expensive to recompute, so the advantage of restoring them is greater (4.25$\times$ / 12.3$\times$, against 2.8$\times$
/ 8.8$\times$ for the 12B). Reuse therefore scales with model size.

\subsection{Deposit (the one-time cost)}\label{deposit-the-one-time-cost}

\begin{center}\footnotesize
\begin{tabular*}{\linewidth}{@{\extracolsep{\fill}}llllll@{}}
\toprule
Model & Tokens & Throughput & Peak VRAM & Encrypted store & Per token \\
\midrule
Gemma 4 12B & 50,000,000 & 20,839 tok/s & 70,864 MB (flat) & 1.86 TB (3,125 $\times$ MRLNCRY1) & 36.4 KiB \\
Gemma 4 31B & 50,000,000 & 10,700 tok/s & 75,111 MB (flat) & 6.25 TB (3,125 $\times$ MRLNCRY1) & 134 KiB \\
\bottomrule
\end{tabular*}
\end{center}

The deposit is a cost that is paid once. Each reuse after that is 2.8$\times$ to 4.3$\times$ cheaper in time and
8.8$\times$ to 12.3$\times$ cheaper in energy than a recompute. During the deposit, VRAM was flat over the whole
50M-token stream, which corresponds to an O(1) moving window.

\subsection{What the numbers say}\label{what-the-numbers-say}

Two observations follow from these results. The first concerns the memory itself, which is exact
and model-agnostic: restore is 100/100 at every depth to 50M on both models, and VRAM is flat
throughout.

The second is that answer quality is determined by the model that does the reading and not by the
memory. On identical stored blocks the 12B reads 82/100 and the 31B 98/100, and neither of them
ever fabricated a code. In every miss the model picked a wrong line that was genuinely in the
block, either a same-format decoy or another in-block number. In practice a cheaper model can
serve and a stronger one can be escalated to, with both free of recompute and both reading the
same durable memory.

\section{Why this is long-term memory and not a cache}\label{why-this-is-long-term-memory-and-not-a-cache}

A cache is short-lived and is rebuilt from the source after a miss. The store we measured differs
from that in four respects.

It is \emph{durable}: the state is read back from disk, it survives a restart, and nothing is pruned or
evicted, so all 3,125 blocks remain (\S5.1). It is \emph{private}, because every block is encrypted at
rest (AES-256-GCM, one key per organisation) and the memory can therefore safely hold an
organisation's or a person's own material. It is \emph{byte-exact}, meaning that a restored block
reproduces the computation and the memory is the model's own understanding of the content and not
a lossy summary. Finally, it reaches \emph{beyond the window}, since the model reuses knowledge from
50M tokens ago, far outside its 16k native window, without recomputing it.

The same mechanism makes inference cheaper (8.8$\times$ to 12.3$\times$ less energy) and faster (2.8$\times$ to 4.3$\times$).
It also makes the outcome smarter, in the specific sense that the model answers from facts it
processed millions of tokens earlier and that are otherwise unavailable to it. The weights are not
changed by any of this; what changes is the set of facts the model can act on. This is the sense
in which we call \texttt{galahad-kv} long-term memory for AI.

\section{Limitations and scope}\label{limitations-and-scope}

\begin{itemize}
\tightlist
\item
  \textbf{Not a wider attention window.} One block is resident, and depth is restored and not attended
  all at once. Processed $\neq$ held-in-VRAM $\neq$ restored.
\item
  \textbf{KV reuse, not learned retrieval.} Selecting a passage by meaning (Blaise) is out of scope
  here, and combining it with this window at 50M scale is left for future work.
\item
  \textbf{Energy resolution.} A per-probe restore takes $\sim$0.3 s, which approaches the resolution of the
  NVML counter. We consider the restore-vs-recompute ratio reliable, but the absolute per-probe
  joules carry a wider band.
\item
  \textbf{Storage.} Durable memory trades disk for reuse (1.86 TB at 12B, 6.25 TB at 31B for 50M), and
  the store has to be on local NVMe, since a network volume voids the latency and energy figures.
\end{itemize}

\section{Reproducibility (single GPU, public software)}\label{reproducibility-single-gpu-public-software}

\begin{Verbatim}[fontsize=\footnotesize,frame=single,framesep=3pt,samepage=true,rulecolor=\color{black!30}]
pip install galahad-kv==1.31.4 vllm==0.31.0 datasets transformers nvidia-ml-py
# obtain a Gemma 4 model; obtain a free Galahad licence on the GPU host; serve with:
#   vllm serve <model> --kv-transfer-config \
#     '{"kv_connector":"GalahadConnector","kv_role":"kv_both"}' \
#     --max-model-len 16640 --enable-prompt-tokens-details
\end{Verbatim}

The four-stage harness (Appendix A) is then run in order: it builds the hardened corpus, deposits
it blind, audits the encrypted store and probes with the two metrics. The harness itself is a thin
public HTTP client (\texttt{transformers}, \texttt{datasets}, \texttt{urllib}) and has no private component, so a third
party may swap in any datasets, model and needle design as long as they follow \S4. For every
reported figure there is a raw artifact and a SHA-256 manifest (Appendix B).

\appendix
\begingroup\raggedright

\section{Harness}\label{harness}

\texttt{build\_hardened.py} (corpus + mutators + needles + decoys + answer key),
\texttt{deposit\_blind.py} (blind ingest, NVML VRAM/energy), \texttt{audit\_storage.py} (MRLNCRY1 + KV-math),
\texttt{probe\_watertight.py} (two metrics, recompute baseline, negative control, miss classification).

\section{Data and manifest}\label{data-and-manifest}

Raw per-depth results: \texttt{RESULT\_12b\_probe.json} (12B), \texttt{RESULT\_31b\_probe.json} (31B); deposit logs;
answer key; \texttt{SHA256SUMS\_31b.txt}. Stores: 1.86 TB (12B), 6.25 TB (31B), 3,125 MRLNCRY1 blocks each,
on local NVMe.

\endgroup

\bigskip
\noindent\emph{Corbenic AI \textperiodcentered{} corbenic.ai \textperiodcentered{} Long-context window: patent pending. \copyright{} 2026 Corbenic AI, Inc.}


\begin{thebibliography}{30}

\small

\bibitem{vaswani2017attention} A. Vaswani, N. Shazeer, N. Parmar, J. Uszkoreit, L. Jones, A. N. Gomez, L. Kaiser, I. Polosukhin. ``Attention Is All You Need.'' NeurIPS, 2017. arXiv:1706.03762.

\bibitem{pope2023scaling} R. Pope, S. Douglas, A. Chowdhery, J. Devlin, J. Bradbury, J. Heek, K. Xiao, S. Agrawal, J. Dean. ``Efficiently Scaling Transformer Inference.'' MLSys, 2023. arXiv:2211.05102.

\bibitem{kwon2023vllm} W. Kwon, Z. Li, S. Zhuang, Y. Sheng, L. Zheng, C. H. Yu, J. E. Gonzalez, H. Zhang, I. Stoica. ``Efficient Memory Management for Large Language Model Serving with PagedAttention.'' SOSP, 2023. arXiv:2309.06180.

\bibitem{gim2024promptcache} I. Gim, G. Chen, S.-s. Lee, N. Sarda, A. Khandelwal, L. Zhong. ``Prompt Cache: Modular Attention Reuse for Low-Latency Inference.'' MLSys, 2024. arXiv:2311.04934.

\bibitem{liu2025lmcache} Y. Liu, Y. Cheng, J. Yao, et al.~``LMCache: An Efficient KV Cache Layer for Enterprise-Scale LLM Inference.'' MLSys, 2026. arXiv:2510.09665.

\bibitem{yao2025cacheblend} J. Yao, H. Li, Y. Liu, S. Ray, Y. Cheng, Q. Zhang, K. Du, S. Lu, J. Jiang. ``CacheBlend: Fast Large Language Model Serving for RAG with Cached Knowledge Fusion.'' EuroSys, 2025. arXiv:2405.16444.

\bibitem{liu2024cachegen} Y. Liu, H. Li, Y. Cheng, et al.~``CacheGen: KV Cache Compression and Streaming for Fast Large Language Model Serving.'' ACM SIGCOMM, 2024. arXiv:2310.07240.

\bibitem{gao2024cachedattention} B. Gao, Z. He, P. Sharma, Q. Kang, D. Jevdjic, J. Deng, X. Yang, Z. Yu, P. Zuo. ``Cost-Efficient Large Language Model Serving for Multi-turn Conversations with CachedAttention.'' USENIX ATC, 2024. arXiv:2403.19708.

\bibitem{lee2024infinigen} W. Lee, J. Lee, J. Seo, J. Sim. ``InfiniGen: Efficient Generative Inference of Large Language Models with Dynamic KV Cache Management.'' USENIX OSDI, 2024. arXiv:2406.19707.

\bibitem{cheng2024cdn} Y. Cheng, K. Du, J. Yao, J. Jiang. ``Do Large Language Models Need a Content Delivery Network?'' 2024. arXiv:2409.13761.

\bibitem{lewis2020rag} P. Lewis, E. Perez, A. Piktus, F. Petroni, V. Karpukhin, N. Goyal, H. K\"uttler, M. Lewis, W.-t. Yih, T. Rockt\"aschel, S. Riedel, D. Kiela. ``Retrieval-Augmented Generation for Knowledge-Intensive NLP Tasks.'' NeurIPS, 2020. arXiv:2005.11401.

\bibitem{chan2025cag} B. J. Chan, C.-T. Chen, J.-H. Cheng, H.-H. Huang. ``Don't Do RAG: When Cache-Augmented Generation is All You Need for Knowledge Tasks.'' Companion Proceedings of the ACM Web Conference, 2025. arXiv:2412.15605.

\bibitem{packer2023memgpt} C. Packer, V. Fang, S. G. Patil, et al.~``MemGPT: Towards LLMs as Operating Systems.'' 2023. arXiv:2310.08560.

\bibitem{gemma4report} Gemma Team, Google DeepMind. ``Gemma 4 Technical Report.'' 2026. arXiv:2607.02770.

\bibitem{gemma3report} Gemma Team, Google DeepMind. ``Gemma 3 Technical Report.'' 2025. arXiv:2503.19786.

\bibitem{gemma4card} Google. ``Gemma 4 model card.'' \url{https://ai.google.dev/gemma/docs/core/model_card_4}. Accessed October 2026.

\bibitem{schelpe2026ceiling} S. Schelpe. ``Working Around the Compute Ceiling: Galahad's Byte-Exact Memory Makes LLM Reading a One-Time Cost.'' 2026. arXiv:2609.39358.

\bibitem{schelpe2026frozen} S. Schelpe. ``A Frozen 12B Beats Frontier Models on Verified Work: 100\% Accuracy, 0 Tokens, Bit-Exact, Forever.'' 2026. arXiv:2607.23806.

\bibitem{schelpe2026flywheel} S. Schelpe. ``Smarter and Cheaper at Once: Byte-Exact KV-State Grafting Turns a Frozen Small Model into a Verified-Knowledge Flywheel.'' 2026. arXiv:2607.14431.

\bibitem{dworkin2007gcm} M. Dworkin. ``Recommendation for Block Cipher Modes of Operation: Galois/Counter Mode (GCM) and GMAC.'' NIST Special Publication 800-38D, 2007. doi:10.6028/NIST.SP.800-38D.

\bibitem{liu2024lost} N. F. Liu, K. Lin, J. Hewitt, A. Paranjape, M. Bevilacqua, F. Petroni, P. Liang. ``Lost in the Middle: How Language Models Use Long Contexts.'' Transactions of the Association for Computational Linguistics, 2024. arXiv:2307.03172.

\bibitem{hsieh2024ruler} C.-P. Hsieh, S. Sun, S. Kriman, S. Acharya, D. Rekesh, F. Jia, Y. Zhang, B. Ginsburg. ``RULER: What's the Real Context Size of Your Long-Context Language Models?'' 2024. arXiv:2404.06654.

\bibitem{golchin2024timetravel} S. Golchin, M. Surdeanu. ``Time Travel in LLMs: Tracing Data Contamination in Large Language Models.'' ICLR, 2024. arXiv:2308.08493.

\bibitem{nvml} NVIDIA. ``NVML API Reference.'' \url{https://docs.nvidia.com/deploy/nvml-api/latest/nvml-api-reference.html}. Accessed October 2026.

\bibitem{wolf2020transformers} T. Wolf, L. Debut, V. Sanh, et al.~``Transformers: State-of-the-Art Natural Language Processing.'' EMNLP: System Demonstrations, 2020. doi:10.18653/v1/2020.emnlp-demos.6.

\bibitem{lhoest2021datasets} Q. Lhoest, A. Villanova del Moral, Y. Jernite, et al.~``Datasets: A Community Library for Natural Language Processing.'' EMNLP: System Demonstrations, 2021. doi:10.18653/v1/2021.emnlp-demo.21.

\bibitem{zhao2024wildchat} W. Zhao, X. Ren, J. Hessel, C. Cardie, Y. Choi, Y. Deng. ``WildChat: 1M ChatGPT Interaction Logs in the Wild.'' ICLR, 2024. arXiv:2405.01470. Dataset: \texttt{allenai/WildChat-1M}.

\bibitem{codeparrot} CodeParrot. Python files from GitHub. Hugging Face dataset \texttt{codeparrot/codeparrot-clean-valid}, \url{https://huggingface.co/datasets/codeparrot/codeparrot-clean-valid}.

\bibitem{cohan2018discourse} A. Cohan, F. Dernoncourt, D. S. Kim, T. Bui, S. Kim, W. Chang, N. Goharian. ``A Discourse-Aware Attention Model for Abstractive Summarization of Long Documents.'' NAACL-HLT, 2018. doi:10.18653/v1/N18-2097. Dataset: \texttt{ccdv/arxiv-summarization}.

\bibitem{penedo2024fineweb} G. Penedo, H. Kydl\'{\i}\v{c}ek, L. Ben Allal, A. Lozhkov, M. Mitchell, C. Raffel, L. Von Werra, T. Wolf. ``The FineWeb Datasets: Decanting the Web for the Finest Text Data at Scale.'' NeurIPS Datasets and Benchmarks Track, 2024. arXiv:2406.17557. Dataset: \texttt{HuggingFaceFW/fineweb-edu} (\texttt{sample-10BT}).

\end{thebibliography}
\end{document}